\documentclass[twoside]{article}
\usepackage[accepted]{aistats2018}
\usepackage[utf8]{inputenc} 
\usepackage[T1]{fontenc}    
\usepackage{hyperref}       
\usepackage{url}            
\usepackage{booktabs}       
\usepackage{amsfonts}       
\usepackage{nicefrac}       
\usepackage{microtype}      

\usepackage{arydshln}
\usepackage{multirow}

\usepackage{widetext}

\usepackage{mathtools, cuted}

\usepackage{floatrow}
\floatstyle{plaintop}
\restylefloat{table}

\usepackage{amsmath}

\usepackage{graphicx}

\graphicspath{{/images}}

\newcommand{\mb}[1]{\mathbf{#1}}

\newcommand{\dd}[0]{\mathrm{d}}

\begin{document}

%

%

\twocolumn[

\aistatstitle{VAE with a VampPrior}

\aistatsauthor{ Jakub M. Tomczak \And Max Welling }

\aistatsaddress{ University of Amsterdam \And University of Amsterdam } ]

\begin{abstract}
Many different methods to train deep generative models have been introduced in the past. In this paper, we propose to extend the variational auto-encoder (VAE) framework with a new type of prior which we call "Variational Mixture of Posteriors" prior, or VampPrior for short. The VampPrior consists of a mixture distribution (\textit{e.g.}, a mixture of Gaussians) with components given by variational posteriors conditioned on learnable pseudo-inputs. We further extend this prior to a two layer hierarchical model and show that this architecture with a coupled prior and posterior, learns significantly better models. The model also avoids the usual local optima issues related to useless latent dimensions that plague VAEs. We provide empirical studies on six datasets, namely, static and binary MNIST, OMNIGLOT, Caltech 101 Silhouettes, Frey Faces and Histopathology patches, and show that applying the hierarchical VampPrior delivers state-of-the-art results on all datasets in the unsupervised permutation invariant setting and the best results or comparable to SOTA methods for the approach with convolutional networks.
\end{abstract}

\section{Introduction}

Learning generative models that are capable of capturing rich distributions from vast amounts of data like image collections remains one of the major challenges of machine learning. In recent years, different approaches to achieving this goal were proposed by formulating alternative training objectives to the log-likelihood \cite{DRG:15, GPMXWOCB:14, LSZ:15} or by utilizing variational inference \cite{BKM:17}. The latter approach could be made especially efficient through the application of the \textit{reparameterization trick} resulting in a highly scalable framework now known as the \textit{variational auto-encoders} (VAE) \cite{KW:13, RMW:14}. Various extensions to deep generative models have been proposed that aim to enrich the variational posterior \cite{DKB:14, NHS:16, RM:15, SKW:15, TW:16, TRB:15}. Recently, it has been noticed that in fact the prior plays a crucial role in mediating between the generative decoder and the variational encoder. Choosing a too simplistic prior like the standard normal distribution could lead to over-regularization and, as a consequence, very poor hidden representations \cite{HJ:16}.

In this paper, we take a closer look at the regularization term of the variational lower bound inspired by the analysis presented in \cite{MSJGF:15}. Re-formulating the variational lower bound gives two regularization terms: the average entropy of the variational posterior, and the cross-entropy between the averaged (over the training data) variational posterior and the prior. The cross-entropy term can be minimized by setting the prior equal to the average of the variational posteriors over training points. However, this would be computationally very expensive. Instead, we propose a new prior that is a \textit{variational mixture of posteriors} prior, or VampPrior for short. Moreover, we present a new two-level VAE that combined with our new prior can learn a very powerful hidden representation.

The contribution of the paper is threefold:
\begin{itemize}
\item We follow the line of research of improving the VAE by making the prior more flexible. We propose a new prior that is a mixture of variational posteriors conditioned on learnable pseudo-data. This allows the variational posterior to learn more a potent latent representation.
\item We propose a new two-layered generative VAE model with two layers of stochastic latent variables based on the VampPrior idea. This architecture effectively avoids the problems of unused latent dimensions. 
\item We show empirically that the VampPrior always outperforms the standard normal prior in different VAE architectures and that the hierarchical VampPrior based VAE achieves state-of-the-art or comparable to SOTA results on six datasets.
\end{itemize}

\section{Variational Auto-Encoder}
Let $\mb{x}$ be a vector of $D$ observable variables and $\mb{z} \in \mathbb{R}^{M}$ a vector of stochastic latent variables. Further, let $p_{\theta}(\mb{x}, \mb{z})$ be a parametric model of the joint distribution. Given data $\mb{X} = \{\mb{x}_1, \ldots, \mb{x}_N\}$ we typically aim at maximizing the average marginal log-likelihood, $\frac{1}{N} \ln p(\mb{X}) = \frac{1}{N}\sum_{i=1}^{N} \ln p(\mb{x}_{i})$, with respect to parameters. However, when the model is parameterized by a neural network (NN), the optimization could be difficult due to the intractability of the marginal likelihood. One possible way of overcoming this issue is to apply \textit{variational inference} and optimize the following lower bound:
\begin{align}\label{eq:elbo}
\mathbb{E}_{\mb{x} \sim q(\mb{x})}[ \ln p(\mb{x}) ] \geq& \mathbb{E}_{\mb{x} \sim q(\mb{x})} \big{[} \mathbb{E}_{q_{\phi}(\mb{z}|\mb{x})}[ \ln p_{\theta}(\mb{x}|\mb{z}) + \nonumber \\
&+ \ln p_{\lambda}(\mb{z})  - \ln q_{\phi}(\mb{z}|\mb{x}) ] \big{]} \\
\stackrel{\Delta}{=}& \mathcal{L}(\phi, \theta, \lambda), \nonumber
\end{align}
where $q(\mb{x}) = \frac{1}{N} \sum_{n=1}^{N} \delta(\mb{x} - \mb{x}_{n})$ is the empirical distribution, $q_{\phi}(\mb{z}|\mb{x})$ is the \textit{variational posterior} (the \textit{encoder}), $p_{\theta}(\mb{x}|\mb{z})$ is the \textit{generative model} (the \textit{decoder}) and $p_{\lambda}(\mb{z})$ is the \textit{prior}, and $\phi, \theta, \lambda$ are their parameters, respectively.

There are various ways of optimizing this lower bound but for continuous $\mb{z}$ this could be done efficiently through the \textit{re-parameterization} of $q_{\phi}(\mb{z}|\mb{x})$ \cite{KW:13, RMW:14}, which yields a \textit{variational auto-encoder} architecture (VAE). Therefore, during learning we consider a Monte Carlo estimate of the second expectation in (\ref{eq:elbo}) using $L$ sample points:
\begin{align}
\widetilde{ \mathcal{L} }(\phi, \theta, \lambda) =& \mathbb{E}_{\mb{x} \sim q(\mb{x})} \Big{[}  \frac{1}{L} \sum_{l=1}^{L} \big{(} \ln p_{\theta}(\mb{x}|\mb{z}_{\phi}^{(l)}) + \label{eq:elbo_MC_1} \\
&+ \ln p_{\lambda}(\mb{z}_{\phi}^{(l)})  - \ln q_{\phi}(\mb{z}_{\phi}^{(l)}|\mb{x}) \big{)} \Big{]} , \label{eq:elbo_MC_2}
\end{align}
where $\mb{z}_{\phi}^{(l)}$ are sampled from $q_\phi(\mb{z}|\mb{x})$ through the re-parameterization trick.

The first component of the objective function can be seen as the expectation of the negative reconstruction error that forces the hidden representation for each data case to be peaked at its specific MAP value. On the contrary, the second and third components constitute a kind of regularization that drives the encoder to match the prior.

We can get more insight into the role of the prior by inspecting the gradient of $\widetilde{ \mathcal{L} }(\phi, \theta, \lambda)$ in (\ref{eq:elbo_MC_1}) and (\ref{eq:elbo_MC_2}) with respect to a single weight $\phi_{i}$ for a single data point $\mb{x}$, see Eq. (\ref{eq:gradient_vae_1}) and (\ref{eq:gradient_vae_2}) in Supplementary Material for details. We notice that the prior plays a role of an ''anchor'' that keeps the posterior close to it, \textit{i.e.}, the term in round brackets in Eq. (\ref{eq:gradient_vae_2}) is $0$ if the posterior matches the prior.

Typically, the encoder is assumed to have a diagonal covariance matrix, \textit{i.e.}, $q_{\phi}(\mb{z}|\mb{x}) = \mathcal{N}\big{(} \mb{z}|\mu_{\phi}(\mb{x}), \mathrm{diag}(\sigma_{\phi}^{2}(\mb{x})) \big{)}$, where $\mu_{\phi}(\mb{x})$ and $\sigma_{\phi}^{2}(\mb{x})$ are parameterized by a NN with weights $\phi$, and the prior is expressed using the standard normal distribution, $p_{\lambda}(\mb{z}) = \mathcal{N}(\mb{z} | \mb{0}, \mb{I})$. The decoder utilizes a suitable distribution for the data under consideration, \textit{e.g.}, the Bernoulli distribution for binary data or the normal distribution for continuous data, and it is parameterized by a NN with weights $\theta$.

\section{The \textit{Variational Mixture of Posteriors} Prior}

\paragraph{Idea} The variational lower-bound consists of two parts, namely, the reconstruction error and the regularization term between the encoder and the prior. However, we can re-write the training objective (\ref{eq:elbo}) to obtain two regularization terms instead of one \cite{MSJGF:15}:
\begin{align}\label{eq:elbo_two_regularizers}
\mathcal{L}(\phi, \theta, \lambda) =& \mathbb{E}_{\mb{x} \sim q(\mb{x})} \big{[} \mathbb{E}_{q_{\phi}(\mb{z}|\mb{x})}[ \ln p_{\theta}(\mb{x}|\mb{z}) ] \big{]} + \\
	&+ \mathbb{E}_{\mb{x} \sim q(\mb{x})} \big{[} \mathbb{H}[q_{\phi}(\mb{z}|\mb{x})] \big{]} + \\
	&- \mathbb{E}_{\mb{z} \sim q(\mb{z}) } [ -\ln p_{\lambda}(\mb{z}) ]
\end{align}
where the first component is the negative reconstruction error, the second component is the expectation of the entropy $\mathbb{H}[\cdot]$ of the variational posterior and the last component is the cross-entropy between the \textit{aggregated posterior} \cite{HJ:16, MSJGF:15}, $q(\mb{z}) = \frac{1}{N} \sum_{n=1}^{N} q_{\phi}(\mb{z}|\mb{x}_{n})$, and the prior. The second term of the objective encourages the encoder to have large entropy (\textit{e.g.}, high variance) for every data case. The last term aims at matching the aggregated posterior and the prior.

Usually, the prior is chosen in advance, \textit{e.g.}, a standard normal prior. However, one could find a prior that optimizes the ELBO by maximizing the following Lagrange function with the Lagrange multiplier $\beta$:
\begin{equation}\label{eq:LagrangeFun}
\max_{p_{\lambda}(\mb{z})} - \mathbb{E}_{\mb{z} \sim q(\mb{z}) } [ -\ln p_{\lambda}(\mb{z}) ] + \beta \Big{(} \int p_{\lambda}(\mb{z}) \dd \mb{z}  - 1 \Big{)}.
\end{equation}
The solution of this problem is simply the aggregated posterior:
\begin{equation}
p_{\lambda}^{*}(\mb{z}) = \frac{1}{N} \sum_{n=1}^{N} q_{\phi}(\mb{z}|\mb{x}_{n}). 
\end{equation}
However, this choice may potentially lead to overfitting \cite{HJ:16, MSJGF:15} and definitely optimizing the recognition model would become very expensive due to the sum over all training points. On the other hand, having a simple prior like the standard normal distribution is known to result in over-regularized models with only few active latent dimensions \cite{BGS:15}. 

In order to overcome issues like overfitting, over-regularization and high computational complexity, the optimal solution, \textit{i.e.}, the aggregated posterior, can be further approximated by a mixture of variational posteriors with \textit{pseudo-inputs}:
\begin{equation}
p_{\lambda}(\mb{z}) = \frac{1}{K} \sum_{k=1}^{K} q_{\phi}(\mb{z}|\mb{u}_{k}),
\end{equation}
where $K$ is the number of pseudo-inputs, and $\mb{u}_{k}$ is a $D$-dimensional vector we refer to as a \textit{pseudo-input}. The pseudo-inputs are learned through backpropagation and can be thought of as hyperparameters of the prior, alongside parameters of the posterior $\phi$, $\lambda = \{\mb{u}_{1}, \ldots, \mb{u}_{K}, \phi\}$. Importantly, the resulting prior is multimodal, thus, it prevents the variational posterior to be over-regularized. On the other hand, incorporating pseudo-inputs prevents from potential overfitting once we pick $K \ll N$, which also makes the model less expensive to train. We refer to this prior as the \textit{variational  mixture of posteriors prior} (VampPrior).

\paragraph{A comparison to a mixture of Gaussians prior} A simpler alternative to the VampPrior  that still approximates the optimal solution of the problem in (\ref{eq:LagrangeFun}) is a mixture of Gaussians (MoG), $p_{\lambda}(\mb{z}) = \frac{1}{K} \sum_{k=1}^{K} \mathcal{N} \big{(} \mu_k, \mathrm{diag}(\sigma_{k}^{2} ) \big{)}$. The hyperparameters of the prior $\lambda = \{\mu_{k}, \mathrm{diag}(\sigma_{k}^{2}) \}_{k=1}^{K}$ are trained by backpropagation similarly to the pseudo-inputs. The MoG prior influences the variational posterior in the same manner to the standard prior and the gradient of the ELBO with respect to the encoder's parameters takes an analogous form to (\ref{eq:gradient_vae_1}) and (\ref{eq:gradient_vae_2}), see Suplementary Material for details.

In the case of the VampPrior, on the other hand, we obtain two advantages over the MoG prior:
\begin{itemize}
\item First, by coupling the prior with the posterior we entertain fewer parameters and the prior and variational posteriors will at all times ``cooperate'' during training.
\item Second, this coupling highly influences the gradient wrt a single weight of the encoder, $\phi_{i}$, for a given $\mb{x}$, see Eq. (\ref{eq:gradient_eep_1}) and (\ref{eq:gradient_eep_2}) in Supplementary Material for details. The differences in (\ref{eq:gradient_eep_1}) and (\ref{eq:gradient_eep_2}) are close to $0$ as long as $q_{\phi}(\mb{z}_{\phi}^{(l)} | \mb{x}) \approx q_{\phi}(\mb{z}_{\phi}^{(l)} | \mb{u}_{k})$. Thus, the gradient is influenced by pseudo-inputs that are \textit{dissimilar} to $\mb{x}$, \textit{i.e.}, if the posterior produces different hidden representations for $\mb{u}_{k}$ and $\mb{x}$. In other words, since this has to hold for every training case, the gradient points towards a solution where the  variational posterior has high variance. On the contrary, the first part of the objective in (\ref{eq:gradient_eep_0}) causes the posteriors to have low variance and map to different latent explanations for each data case. These effects distinguish the VampPrior from the MoG prior utilized in the VAE so far \cite{DMGLSAS:16, NHS:16}.
\end{itemize}

\paragraph{A connection to the \textit{Empirical Bayes}} The idea of the Empirical Bayes (EB), also known as the type-II maximum likelihood, is to find hyperparameters $\lambda$ of the prior over latent variables $\mb{z}$, $p(\mb{z}|\lambda)$, by maximizing the marginal likelihood $p(\mb{x}|\lambda)$. In the case of the VAE and the VampPrior the pseudo-inputs, alongside the parameters of the posterior, are the hyperparameters of the prior and we aim at maximizing the ELBO with respect to them. Thus, our approach is closely related to the EB and in fact it formulates a new kind of Bayesian inference that combines the variational inference with the EB approach.

\paragraph{A connection to the \textit{Information Bottleneck}} We have shown that the aggregated posterior is the optimal prior within the VAE formulation. This result is closely related to the \textit{Information Bottleneck} (IB) approach \cite{AFDM:17, TPB:00} where the aggregated posterior naturally plays the role of the prior. Interestingly, the VampPrior brings the VAE and the IB formulations together and highlights their close relation. A similar conclusion and a more thorough analysis of the close relation between the VAE and the IB through the VampPrior is presented in \cite{APFDSM:17}.

\begin{figure*}[!htbp]
\includegraphics[width=0.425\textwidth]{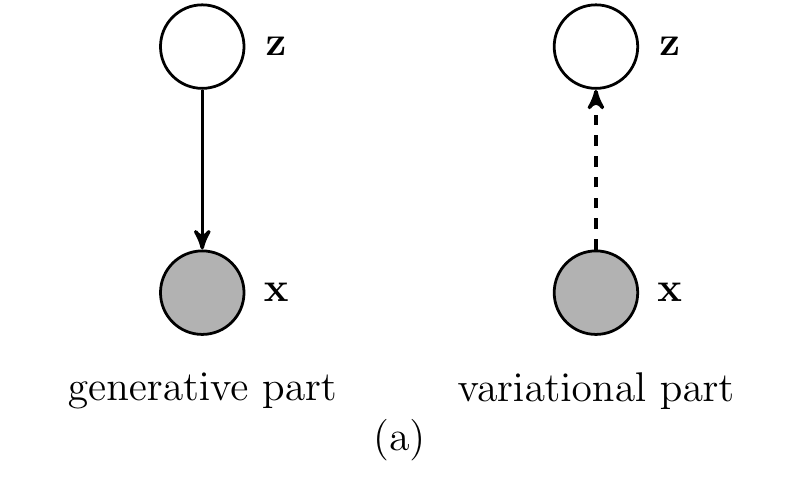}
\qquad
\includegraphics[width=0.425\textwidth]{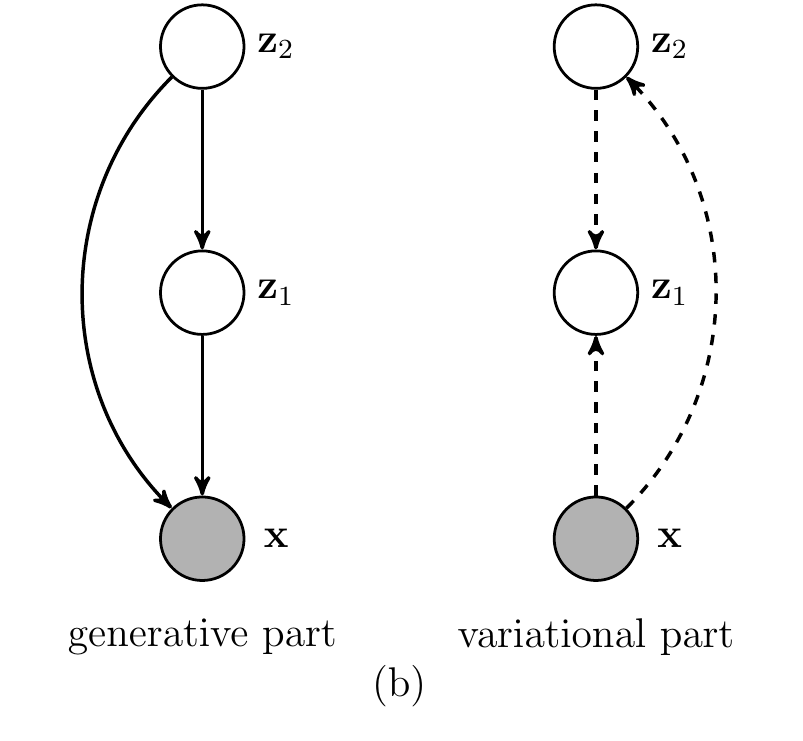}
\vskip -0.25cm
\caption{Stochastical dependencies in: (a) a one-layered VAE and (b) a two-layered model. The generative part is denoted by the solid line and the variational part is denoted by the dashed line.}
\label{fig:models}
\end{figure*}

\section{Hierarchical VampPrior Variational Auto-Encoder}

\paragraph{Hierarchical VAE and the \textit{inactive stochastic latent variable problem}} A typical problem encountered during training a VAE is the \textit{inactive stochastic units} \cite{BGS:15, MFW:17}. Our VampPrior VAE seems to be an effective remedy against this issue, simply because the prior is designed to be rich and multimodal, preventing the KL term from pulling individual posteriors towards a simple (\textit{e.g.}, standard normal) prior. The \textit{inactive stochastic units problem} is even worse for learning deeper VAEs (\textit{i.e.}, with multiple layers of stochastic units). The reason might be that stochastic dependencies within a deep generative model are top-down in the generative process and bottom-up in the variational process. As a result, there is less information obtained from real data at the deeper stochastic layers, making them more prone to become regularized towards the prior. 

In order to prevent a deep generative model to suffer from the inactive stochastic units problem we propose a new two-layered VAE as follows:
\begin{equation}\label{eq:dgm_variational}
q_{\phi}(\mb{z}_1 | \mb{x}, \mb{z}_{2})\ q_{\psi}(\mb{z}_{2} | \mb{x}) ,
\end{equation}
while the the generative part is the following:
\begin{equation}\label{eq:dgm_generative}
p_{\theta}(\mb{x} | \mb{z}_{1}, \mb{z}_{2})\ p_{\lambda}(\mb{z}_1 | \mb{z}_{2})\  p(\mb{z}_{2}) ,
\end{equation}
with $p(\mb{z}_{2})$ given by a VampPrior. 
The model is depicted in Figure \ref{fig:models}(b).

In this model we use normal distributions with diagonal covariance matrices to model $\mb{z}_1 \in \mathbb{R}^{M_1}$ and $\mb{z}_2 \in \mathbb{R}^{M_2}$, parameterized by NNs. The full model is given by:
\begin{align}
p(\mb{z}_{2}) &= \frac{1}{K} \sum_{k=1}^{K} q_{\psi}(\mb{z}_{2} | \mb{u}_{k}) ,\\
p_{\lambda}(\mb{z}_1 | \mb{z}_{2}) &= \mathcal{N}\big{(} \mb{z}_1|\mu_{\lambda}(\mb{z}_2), \mathrm{diag}(\sigma_{\lambda}^{2}(\mb{z}_2)) \big{)} , \\
q_{\phi}(\mb{z}_1 | \mb{x}, \mb{z}_{2}) &= \mathcal{N}\big{(} \mb{z}_1|\mu_{\phi}(\mb{x}, \mb{z}_{2}), \mathrm{diag}(\sigma_{\phi}^{2}(\mb{x}, \mb{z}_{2})) \big{)} , \\
q_{\psi}(\mb{z}_2 | \mb{x}) &= \mathcal{N}\big{(} \mb{z}_2|\mu_{\psi}(\mb{x}), \mathrm{diag}(\sigma_{\psi}^{2}(\mb{x})) \big{)} .
\end{align}

\paragraph{Alternative priors} We have motivated the VampPrior by analyzing the variational lower bound. However, one could inquire whether we really need such complicated prior and maybe the proposed two-layered VAE is already sufficiently powerful. In order to answer these questions we further verify three alternative priors:
\begin{itemize}
\item the standard Gaussian prior (SG): $$p(\mb{z}_{2}) = \mathcal{N}(\mb{0}, \mb{I});$$
\item the mixture of Gaussians prior (MoG): $$p(\mb{z}_{2}) = \frac{1}{K} \sum_{k=1}^{K} \mathcal{N} \big{(} \mu_k, \mathrm{diag}(\sigma_{k}^{2} ) \big{)},$$ where $\mu_{k} \in \mathbb{R}^{M_2}$, $\sigma_{k}^{2} \in \mathbb{R}^{M_2}$ are trainable parameters;
\item the VampPrior with a random subset of real training data as non-trainable pseudo-inputs (VampPrior \textit{data}).
\end{itemize}
Including the standard prior can provide us with an answer to the general question if there is even a need for complex priors. Utilizing the mixture of Gaussians verifies whether it is beneficial to couple the prior with the variational posterior or not. Finally, using a subset of real training images determines to what extent it is useful to introduce trainable pseudo-inputs.

\section{Experiments}

\subsection{Setup}

In the experiments we aim at: (i) verifying empirically whether the VampPrior helps the VAE to train a representation that better reflects variations in data, and (ii) inspecting if our proposition of a two-level generative model performs better than the one-layered model. In order to answer these questions we compare different models parameterized by feed-forward neural networks (MLPs) or convolutional networks that utilize the standard prior and the VampPrior. In order to compare the hierarchical VampPrior VAE with the state-of-the-art approaches, we used also an autoregressive decoder. Nevertheless, our primary goal is to quantitatively and qualitatively assess the newly proposed prior.

We carry out experiments using six image datasets: static MNIST \cite{LM:11}, dynamic MNIST \cite{SM:08}, OMNIGLOT \cite{LST:15}, Caltech 101 Silhouette \cite{MSCF:10}, Frey Faces\footnote{\url{http://www.cs.nyu.edu/~roweis/data/frey_rawface.mat}} and Histopathology patches \cite{TW:16}. More details about the datasets can be found in the Supplementary Material.

In the experiments we modeled all distributions using MLPs with two hidden layers of $300$ hidden units in the unsupervised permutation invariant setting. We utilized the gating mechanism as an element-wise non-linearity \cite{DFAG:16}. We utilized $40$ stochastic hidden units for both $\mb{z}_1$ and $\mb{z}_2$. Next we replaced MLPs with convolutional layers with gating mechanism. Eventually, we verified also a PixelCNN \cite{OKK:16} as the decoder. For Frey Faces and Histopathology we used the discretized logistic distribution of images as in \cite{KSJCSW:16}, and for other datasets we applied the Bernoulli distribution.

For learning the ADAM algorithm \cite{KB:14} with normalized gradients \cite{YWQSC:17} was utilized with the learning rate in $\{10^{-4},5\cdot 10^{-4}\}$ and mini-batches of size $100$. Additionally, to boost the generative capabilities of the decoder, we used the \textit{warm-up} for $100$ epochs \cite{BVDJB:16}. The weights of the neural networks were initialized according to \cite{GB:10}. The early-stopping with a look ahead of $50$ iterations was applied. For the VampPrior we used $500$ pseudo-inputs for all datasets except OMNIGLOT for which we utilized $1000$ pseudo-inputs. For the VampPrior \textit{data} we randomly picked training images instead of the learnable pseudo-inputs.

We denote the hierarchical VAE proposed in this paper with MLPs by \textsc{HVAE}, while the hierarchical VAE with convolutional layers and additionally with a PixelCNN decoder are denoted by \textsc{convHVAE} and \textsc{PixelHVAE}, respectively.

%


\subsection{Results}

\begin{table*}[t]
\caption{Test log-likelihood (LL) between different models with the standard normal prior (standard) and the VampPrior. For last two datasets \textit{an average of bits per data dimension} is given. In the case of Frey Faces, for all models the standard deviation was no larger than $0.03$ and that this why it is omitted in the table.}
\label{tab:comparison}
{\def\arraystretch{1.25}
\begin{tabular}{c|cc|cc|cc|cc|cc}
 & \multicolumn{2}{c|}{\footnotesize{\textsc{VAE} ($L=1$)}} & \multicolumn{2}{c|}{\footnotesize{\textsc{HVAE} ($L=2$)}} & \multicolumn{2}{c|}{\footnotesize{\textsc{convHVAE} ($L=2$)}} & \multicolumn{2}{c|}{\footnotesize{\textsc{PixelHVAE} ($L=2$)}}\\
\textsc{Dataset} & {\fontsize{7}{8.4}\selectfont standard} & {\fontsize{7}{8.4}\selectfont VampPrior} & {\fontsize{7}{8.4}\selectfont standard} & {\fontsize{7}{8.4}\selectfont VampPrior} & {\fontsize{7}{8.4}\selectfont standard} & {\fontsize{7}{8.4}\selectfont VampPrior} & {\fontsize{7}{8.4}\selectfont standard} & {\fontsize{7}{8.4}\selectfont VampPrior} \\
\hline
{\fontsize{7}{8.4}\selectfont staticMNIST } & $-88.56$ & $\mb{-85.57}$ & $-86.05$ & $\mb{-83.19}$ & $-82.41$ & $\mb{-81.09}$ & $-80.58$ & $\mb{-79.78}$ \\
{\fontsize{7}{8.4}\selectfont dynamicMNIST } & $-84.50$ & $\mb{-82.38}$ & $-82.42$ & $\mb{-81.24}$ & $-80.40$ & $\mb{-79.75}$ & $-79.70$ & $\mb{-78.45}$ \\
{\fontsize{7}{8.4}\selectfont Omniglot } & $-108.50$ & $\mb{-104.75}$ & $-103.52$ & $\mb{-101.18}$ & $-97.65$ & $\mb{-97.56}$ & $-90.11$ & $\mb{-89.76}$ \\
{\fontsize{7}{8.4}\selectfont Caltech 101} & $-123.43$ & $\mathbf{-114.55}$ & $-112.08$ & $\mb{-108.28}$ & $-106.35$ & $\mb{-104.22}$ & $\mb{-85.51}$ & $-86.22$ \\
{\fontsize{7}{8.4}\selectfont Frey Faces } & $4.63$ & $\mb{4.57}$ & $4.61$ & $\mb{4.51}$ & $4.49$ & $\mb{4.45}$ & $4.43$ & $\mb{4.38}$ \\
{\fontsize{7}{8.4}\selectfont Histopathology } & $6.07$ & $\mb{6.04}$ & $5.82$ & $\mb{5.75}$ & $5.59$ & $\mb{5.58}$ & $4.84$ & $\mb{4.82}$ 
\end{tabular}
}
\end{table*}

\paragraph{Quantitative results} We quantitatively evaluate our method using the test marginal log-likelihood (LL) estimated using the Importance Sampling with 5,000 sample points \cite{BGS:15, RMW:14}. In Table \ref{tab:comparison} we present a comparison between models with the standard prior and the VampPrior. The results of our approach in comparison to the state-of-the-art methods are gathered in Table \ref{tab:static_mnist}, \ref{tab:dynamic_mnist}, \ref{tab:omniglot} and \ref{tab:caltech} for static and dynamic MNIST, OMNIGLOT and Caltech 101 Silhouettes, respectively.

\begin{table}
\ttabbox
  {\def\arraystretch{1}
  \begin{tabular}{lc}
    \textsc{Model} & \textsc{LL} \\
    \hline   
    \textsc{VAE ($L=1$) + NF} \cite{RM:15} & $-85.10$ \\
    \textsc{VAE ($L=2$)} \cite{BGS:15} & $-87.86$ \\
    \textsc{IWAE ($L=2$)} \cite{BGS:15} & $-85.32$ \\
    \textsc{HVAE ($L=2$) + SG} & $-85.89$ \\
    \textsc{HVAE ($L=2$) + MoG} & $-85.07$ \\
    \footnotesize{\textsc{HVAE ($L=2$) + VampPrior} \textit{data}} & $-85.71$\\
    \textsc{HVAE ($L=2$) + VampPrior} & $\mb{-83.19}$\\
    \hline
    \hline
	\textsc{AVB + AC ($L=1$)} \cite{MNG:17} & $-80.20$\\    
    \textsc{VLAE} \cite{CKSDDSSA:16} & $\mb{-79.03}$\\
    \textsc{VAE + IAF} \cite{KSJCSW:16} & $-79.88$ \\
    \footnotesize{\textsc{convHVAE ($L=2$) + VampPrior}} & $-81.09$\\
    \footnotesize{\textsc{PixelHVAE ($L=2$) + VampPrior}} & $-79.78$\\
  \end{tabular}
  \vspace{0.0mm}
  }
  {\caption{Test LL for static MNIST.}
  \label{tab:static_mnist}}
\end{table}

\begin{table}
{\def\arraystretch{1}
  \begin{tabular}{lc}
    \textsc{Model} & \textsc{LL} \\
    \hline
    \textsc{VAE ($L=2$) + VGP} \cite{TRB:15} & $-81.32$\\
    \textsc{CaGeM-0 ($L=2$) \cite{MFW:17}} & $-81.60$\\
    \textsc{LVAE ($L=5$) \cite{SKTSKW:16}} & $-81.74$ \\
    \footnotesize{\textsc{HVAE ($L=2$) + VampPrior} \textit{data}} & $-81.71$\\
    \textsc{HVAE ($L=2$) + VampPrior} & $\mb{-81.24}$\\
    \hline
    \hline
    \textsc{VLAE} \cite{CKSDDSSA:16} & $-78.53$\\
    \textsc{VAE + IAF} \cite{KSJCSW:16} & $-79.10$ \\
    \textsc{PixelVAE} \cite{GKATVVC:16} & $-78.96$\\
    \footnotesize{\textsc{convHVAE ($L=2$) + VampPrior}} & $-79.78$\\
    \footnotesize{\textsc{PixelHVAE ($L=2$) + VampPrior}} & $\mb{-78.45}$\\
  \end{tabular}
  \vspace{0.0mm}
  }
  {\caption{Test LL for dynamic MNIST.}
  \label{tab:dynamic_mnist}}
\end{table}

\begin{table}
\ttabbox
  {\def\arraystretch{1}
  \begin{tabular}{lc}
    \textsc{Model} & \textsc{LL} \\
    \hline
    \textsc{VR-max ($L=2$)} \cite{LT:16} & $-103.72$ \\
    \textsc{IWAE ($L=2$)} \cite{BGS:15} & $-103.38$ \\
    \textsc{LVAE ($L=5$)} \cite{SKTSKW:16} & $-102.11$ \\
    \textsc{HVAE ($L=2$) + VampPrior} & $\mb{-101.18}$\\
    \hline
    \hline
    \textsc{VLAE} \cite{CKSDDSSA:16} & $-89.83$\\
    \footnotesize{\textsc{convHVAE ($L=2$) + VampPrior}} & $-97.56$\\
    \footnotesize{\textsc{PixelHVAE ($L=2$) + VampPrior}} & $\mb{-89.76}$\\
  \end{tabular}
  \vspace{0.0mm}
  }
  {\caption{Test LL for OMNIGLOT.}
  \label{tab:omniglot}}
\end{table}

\begin{table}
{\def\arraystretch{1.25}
  \begin{tabular}{lc}
    \textsc{Model} & \textsc{LL} \\
    \hline
    \textsc{IWAE ($L=1$)} \cite{LT:16} & $-117.21$ \\
    \textsc{VR-max ($L=1$)} \cite{LT:16} & $-117.10$ \\
    \textsc{HVAE ($L=2$) + VampPrior} & $\mb{-108.28}$\\
    \hline
    \hline
    \textsc{VLAE} \cite{CKSDDSSA:16} & $\mb{-78.53}$\\
    \footnotesize{\textsc{convHVAE ($L=2$) + VampPrior}} & $-104.22$\\
    \footnotesize{\textsc{PixelHVAE ($L=2$) + VampPrior}} & $-86.22$\\
  \end{tabular}
  \vspace{0.0mm}
  }
  {\caption{Test LL for Caltech 101 Silhouettes.}
  \label{tab:caltech}}
\end{table}

First we notice that in all cases except one the application of the VampPrior results in a substantial improvement of the generative performance in terms of the test LL comparing to the standard normal prior (see Table \ref{tab:comparison}). This confirms our supposition that a combination of multimodality and coupling the prior with the posterior is superior to the standard normal prior. Further, we want to stress out that the VampPrior outperforms other priors like a single Gaussian or a mixture of Gaussians (see Table \ref{tab:static_mnist}). These results provide an additional evidence that the VampPrior leads to a more powerful latent representation and it differs from the MoG prior. We also examined whether the presented two-layered model performs better than the widely used hierarchical architecture of the VAE. Indeed, the newly proposed approach is more powerful even with the SG prior (HVAE ($L=2$) + SG) than the standard two-layered VAE (see Table \ref{tab:static_mnist}). Applying the MoG prior results in an additional boost of performance. This provides evidence for the usefulness of a multimodal prior. The VampPrior \textit{data} gives only slight improvement comparing to the SG prior and because of the application of the fixed training data as the pseudo-inputs it is less flexible than the MoG. Eventually, coupling the variational posterior with the prior and introducing learnable pseudo-inputs gives the best performance.

\begin{figure}[!htbp]
\includegraphics[width=1.0\textwidth]{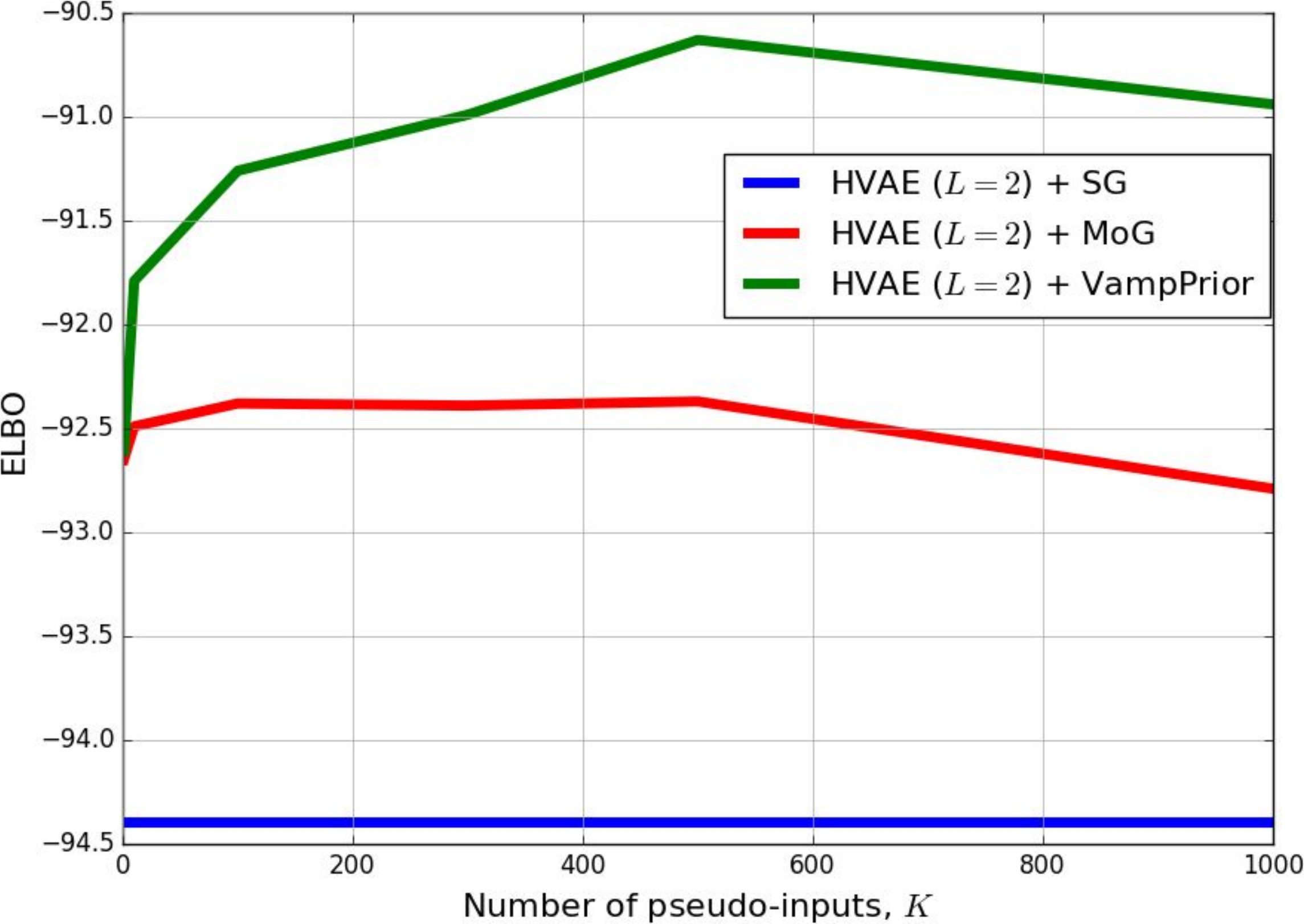}
\vskip 0.2cm
\caption{A comparison between the HVAE ($L=2$) with SG prior, MoG prior and VampPrior in terms of ELBO and varying number of pseudo-inputs/components on static MNIST.}
\label{fig:comparison_a}
\end{figure}

Additionally, we compared the VampPrior with the MoG prior and the SoG prior in Figure \ref{fig:comparison_a} for varying number of pseudo-inputs/components. Surprisingly, taking more pseudo-inputs does not help to improve the performance and, similarly, considering more mixture components also resulted in drop of the performance. However, again we can notice that the VampPrior is more flexible prior that outperforms the MoG.

An inspection of histograms of the log-likelihoods (see Supplementary Material) shows that the distributions of LL values are heavy-tailed and/or bimodal. A possible explanation for such characteristics of the histograms is the existence of many examples that are relatively simple to represent (first mode) and some really \textit{hard} examples (heavy-tail). Comparing our approach to the VAE reveals that the VAE with the VampPrior is not only better on average but it produces less examples with high values of LL and more examples with lower LL.

\begin{figure}[!htbp]
\includegraphics[width=1.0\textwidth]{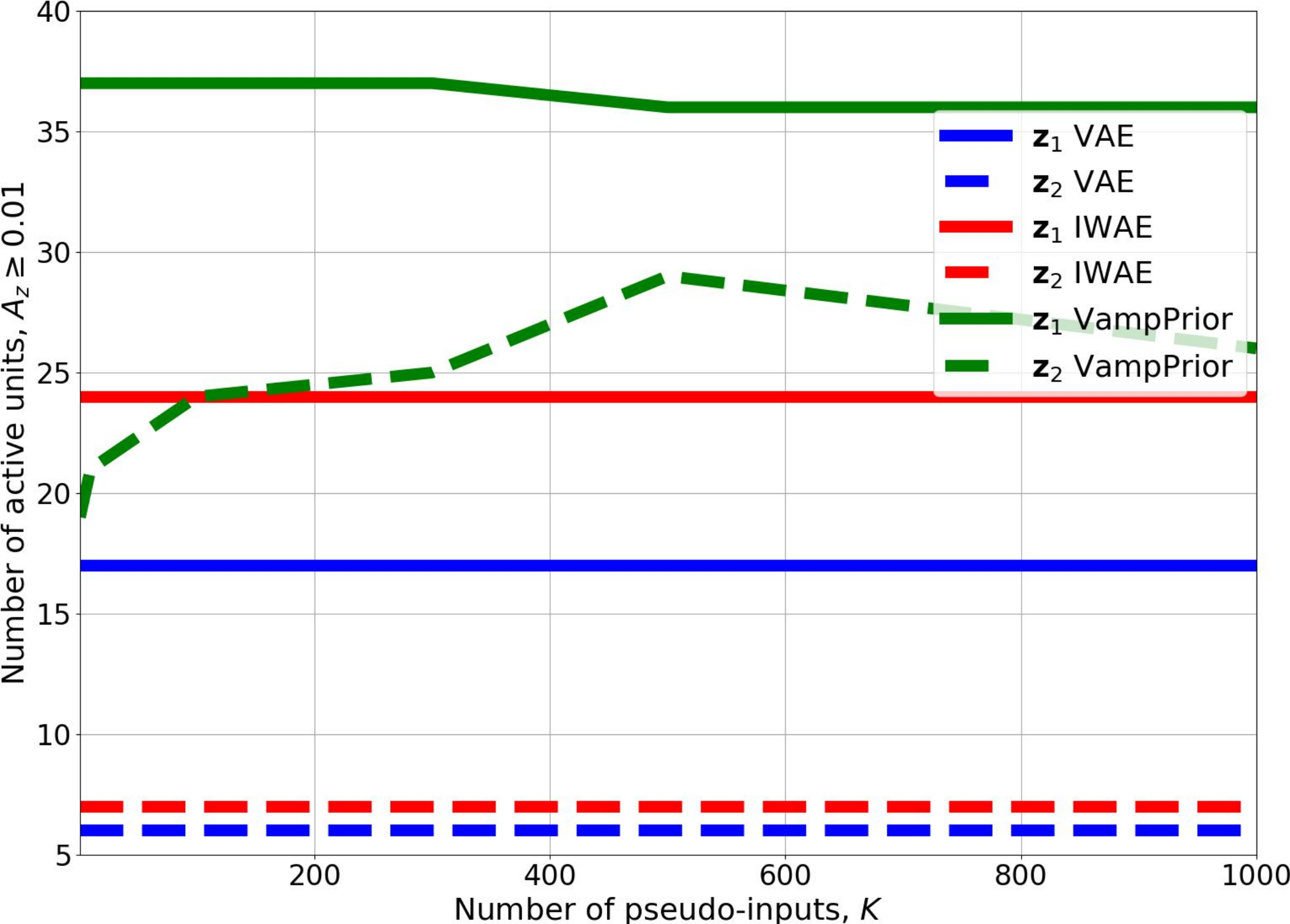}
\vskip 0.2cm
\caption{A comparison between two-level VAE and IWAE with the standard normal prior and theirs VampPrior counterpart in terms of number of active units for varying number of pseudo-inputs on static MNIST.}
\label{fig:comparison_b}
\end{figure}

We hypothesized that the VampPrior provides a remedy for the inactive units issue. In order to verify this claim we utilized the statistics introduced in \cite{BGS:15}. The results for the HVAE with the VampPrior in comparison to the two-level VAE and IWAE presented in \cite{BGS:15} are given in Figure \ref{fig:comparison_b}. The application of the VampPrior increases the number of active stochastic units four times for the second level and around 1.5 times more for the first level comparing to the VAE and the IWAE. Interestingly, the number of mixture components has a great impact on the number of active stochastic units in the second level. Nevertheless, even one mixture component allows to achieve almost three times more active stochastic units comparing to the vanilla VAE and the IWAE.

In general, an application of the VampPrior improves the performance of the VAE and in the case of two layers of stochastic units it yields the state-of-the-art results on all datasets for models that use MLPs. Moreover, our approach gets closer to the performance of models that utilize convolutional neural networks, such as, the one-layered VAE with the inverse autoregressive flow (IAF) \cite{KSJCSW:16} that achieves $-79.88$ on static MNIST and $-79.10$ on dynamic MNIST, the one-layered Variational Lossy Autoencoder (VLAE) \cite{CKSDDSSA:16} that obtains $-79.03$ on static MNIST and $-78.53$ on dynamic MNIST, and the hierarchical PixelVAE \cite{GKATVVC:16} that gets $-78.96$ on dynamic MNIST. On the other two datasets the VLAE performs way better than our approach with MLPs and achieves $-89.83$ on OMNIGLOT and $-77.36$ on Caltech 101 Silhouettes.

In order to compare our approach to the state-of-the-art convolutional-based VAEs we performed additional experiments using the HVAE ($L=2$) + VampPrior with convolutional layers in both the encoder and decoder. Next, we replaced the convolutional decoder with the PixelCNN \cite{OKK:16} decoder (\textsc{convHVAE} and \textsc{PixelHVAE} in Tables \ref{tab:static_mnist}--\ref{tab:caltech}). For the \textsc{PixelHVAE} we were able to improve the performance to obtain $-79.78$ on static MNIST, $-78.45$ on dynamic MNIST, $-89.76$ on Omniglot, and $-86.22$ on Caltech 101 Silhouettes. The results confirmed that the VampPrior combined with a powerful encoder and a flexible decoder performs much better than the MLP-based approach and allows to achieve state-of-the-art performance on dynamic MNIST and OMNIGLOT\footnote{In \cite{CKSDDSSA:16} the performance of the VLAE on static MNIST and Caltech 101 Silhouettes is provided for a different training procedure than the one used in this paper.}.

\paragraph{Qualitative results} The biggest disadvantage of the VAE is that it tends to produce blurry images \cite{LSLW:16}. We noticed this effect in images generated and reconstructed by VAE (see Supplementary Material). Moreover, the standard VAE produced some digits that are hard to interpret, blurry characters and very noisy silhouettes. The supremacy of HVAE + VampPrior is visible not only in LL values but in image generations and reconstructions as well because these are sharper.

\begin{figure*}[!htbp]
\includegraphics[width=1.0\textwidth]{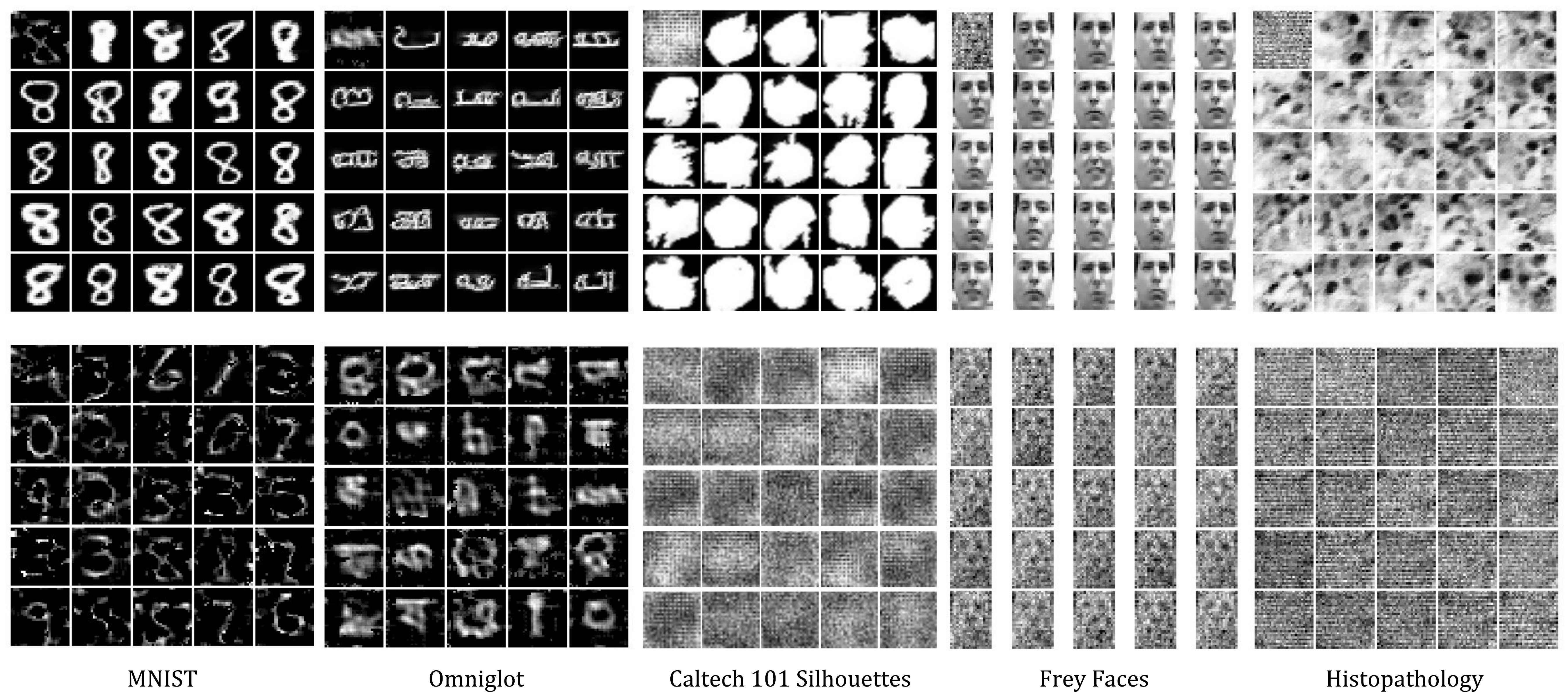}
\vskip -0.0cm
\caption{(\textit{top row}) Images generated by \textsc{PixelHVAE + VampPrior} for chosen pseudo-input in the left top corner. (\textit{bottom row}) Images represent a subset of trained pseudo-inputs for different datasets.}
\label{fig:VampPrior_gen}
\end{figure*}

We also examine what the pseudo-inputs represent at the end of the training process (see Figure \ref{fig:VampPrior_gen}). Interestingly, trained pseudo-inputs are prototypical objects (digits, characters, silhouettes). Moreover, images generated for a chosen pseudo-input show that the model encodes a high variety of different features such as shapes, thickness and curvature for a single pseudo-input. This means that the model is not just memorizing the data-cases. It is worth noticing that for small-sample size and/or too flexible decoder the pseudo-inputs can be hard to train and they can represent noisy prototypes (\textit{e.g.}, see pseudo-inputs for Frey Faces in Figure \ref{fig:VampPrior_gen}).

\section{Related work}

The VAE is a latent variable model that is usually trained with a very simple prior, \textit{i.e.}, the standard normal prior. In \cite{NS:16} a Dirichlet process prior using a stick-breaking process was proposed, while \cite{GHLWX:17} proposed a nested Chinese Restaurant Process. These priors enrich the generative capabilities of the VAE, however, they require sophisticated learning methods and tricks to be trained successfully. A different approach is to use an autoregressive prior \cite{CKSDDSSA:16} that applies the IAF to random noise. This approach gives very promising results and allows to build rich representations. Nevertheless, the authors of \cite{CKSDDSSA:16} combine their prior with a convolutional encoder and an autoregressive decoder that makes it harder to assess the real contribution of the autoregressive prior to the generative model.

Clearly, the quality of generated images are dependent on the decoder architecture. One way of improving generative capabilities of the decoder is to use an infinite mixture of probabilistic component analyzers \cite{SC:16}, which is equivalent to a rank-one covariance matrix. A more appealing approach would be to use deep autoregressive density estimators that utilize recurrent neural networks \cite{OKK:16} or gated convolutional networks \cite{OKEVG:16}. However, there is a threat that a too flexible decoder could discard hidden representations completely, turning the encoder to be useless \cite{CKSDDSSA:16}. 


Concurrently to our work, in \cite{BMZR:17} a variational VAE with a memory was proposed. This approach shares similarities with the VampPrior in terms of a learnable memory (pseudo-inputs in our case) and a multimodal prior. Nevertheless, there are two main differences. First, our prior is an explicit mixture while they sample components. Second, we showed that the optimal prior requires to be coupled with the variational posterior. In the experiments we showed that the VampPrior improves the generative capabilities of the VAE but in \cite{BMZR:17} it was noticed that the generative performance is comparable to the standard normal prior. We claim that the success of the VampPrior follows from the utilization of the variational posterior in the prior, a part that is missing in \cite{BMZR:17}.

Very recently, the VampPrior was shown to be a perfect match for the information-theoretic approach to learn latent representation \cite{APFDSM:17}. Additionally, the authors of \cite{APFDSM:17} proposed to use a weighted version of the VampPrior:
\begin{equation}
p_{\lambda}(\mb{z}) = \sum_{k=1}^{K} w_{k} q_{\phi}(\mb{z}|\mb{u}_{k}),
\end{equation}
where $w_1, \ldots, w_K$ are trainable parameters such that $ \forall_k w_k \geq 0$ and $\sum_k w_k = 1$. This allows the VampPrior to learn which components (\textit{i.e}, pseudo-inputs) are meaningful and may prevent from potential overfitting.

\section{Conclusion}
In this paper, we followed the line of thinking that the prior is a critical element to improve deep generative models, and in particular VAEs. We proposed a new prior that is expressed as a mixture of variational posteriors. In order to limit the capacity of the prior we introduced learnable pseudo-inputs as hyper-parameters of the prior, the number of which can be chosen freely. Further, we formulated a new two-level generative model based on this VampPrior. We showed empirically that applying our prior can indeed increase the performance of the proposed generative model and successfully overcome the problem of inactive stochastic latent variables, which is particularly challenging for generative models with multiple layers of stochastic latent variables. As a result, we achieved state-of-the-art or comparable results to SOTA models on six datasets. Additionally, generations and reconstructions obtained from the hierarchical VampPrior VAE are of better quality than the ones achieved by the standard VAE.

We believe that it is worthwhile to further pursue the line of the research presented in this paper. Here we applied our prior to image data but it would be interesting to see how it behaves on text or sound, where the sequential aspect plays a crucial role. We have already showed that combining the VampPrior VAE with convolutional nets and a powerful autoregressive density estimator is beneficial but more thorough study is needed. Last but not least, it would be interesting to utilize a normalizing flow \cite{RM:15}, the hierarchical variational inference \cite{RTB:16}, ladder networks \cite{SKTSKW:16} or adversarial training \cite{MNG:17} within the VampPrior VAE. However, we leave investigating these issues for future work.

\subsubsection*{Acknowledgements}
The authors are grateful to Rianne van den Berg, Yingzhen Li, Tim Genewein, Tameem Adel, Ben Poole and Alex Alemi for insightful comments.

The research conducted by Jakub M. Tomczak was funded by the European Commission within the Marie Sk\l odowska-Curie Individual Fellowship (Grant No. 702666, ''Deep learning and Bayesian inference for medical imaging'').


\begin{thebibliography}{10}

\bibitem{AFDM:17}
A.~Alemi, I.~Fischer, J.~Dillon, and K.~Murphy.
\newblock Deep variational information bottleneck.
\newblock {\em ICLR}, 2017.

\bibitem{APFDSM:17}
A.~A. Alemi, B.~Poole, I.~Fischer, J.~V. Dillon, R.~A. Saurous, and K.~Murphy.
\newblock An information-theoretic analysis of deep latent-variable models.
\newblock {\em arXiv preprint arXiv:1711.00464}, 2017.

\bibitem{BKM:17}
D.~M. Blei, A.~Kucukelbir, and J.~D. McAuliffe.
\newblock Variational inference: A review for statisticians.
\newblock {\em Journal of the American Statistical Association}, 2017.

\bibitem{BMZR:17}
J.~Bornschein, A.~Mnih, D.~Zoran, and D.~J. Rezende.
\newblock {Variational Memory Addressing in Generative Models}.
\newblock {\em arXiv:1709.07116}, 2017.

\bibitem{BVDJB:16}
S.~R. Bowman, L.~Vilnis, O.~Vinyals, A.~M. Dai, R.~Jozefowicz, and S.~Bengio.
\newblock Generating sentences from a continuous space.
\newblock {\em arXiv:1511.06349}, 2016.

\bibitem{BGS:15}
Y.~Burda, R.~Grosse, and R.~Salakhutdinov.
\newblock Importance weighted autoencoders.
\newblock {\em arXiv:1509.00519}, 2015.

\bibitem{CKSDDSSA:16}
X.~Chen, D.~P. Kingma, T.~Salimans, Y.~Duan, P.~Dhariwal, J.~Schulman,
  I.~Sutskever, and P.~Abbeel.
\newblock {Variational Lossy Autoencoder}.
\newblock {\em arXiv:1611.02731}, 2016.

\bibitem{DFAG:16}
Y.~N. Dauphin, A.~Fan, M.~Auli, and D.~Grangier.
\newblock {Language Modeling with Gated Convolutional Networks}.
\newblock {\em arXiv:1612.08083}, 2016.

\bibitem{DMGLSAS:16}
N.~Dilokthanakul, P.~A. Mediano, M.~Garnelo, M.~C. Lee, H.~Salimbeni,
  K.~Arulkumaran, and M.~Shanahan.
\newblock Deep unsupervised clustering with gaussian mixture variational
  autoencoders.
\newblock {\em arXiv:1611.02648}, 2016.

\bibitem{DKB:14}
L.~Dinh, D.~Krueger, and Y.~Bengio.
\newblock {NICE: Non-linear independent components estimation}.
\newblock {\em arXiv:1410.8516}, 2014.

\bibitem{DRG:15}
G.~Dziugaite, D.~Roy, and Z.~Ghahramani.
\newblock Training generative neural networks via maximum mean discrepancy
  optimization.
\newblock {\em UAI}, pages 258--267, 2015.

\bibitem{GB:10}
X.~Glorot and Y.~Bengio.
\newblock Understanding the difficulty of training deep feedforward neural
  networks.
\newblock {\em AISTATS}, 9:249--256, 2010.

\bibitem{GPMXWOCB:14}
I.~Goodfellow, J.~Pouget-Abadie, M.~Mirza, B.~Xu, D.~Warde-Farley, S.~Ozair,
  A.~Courville, and Y.~Bengio.
\newblock Generative adversarial nets.
\newblock {\em NIPS}, pages 2672--2680, 2014.

\bibitem{GHLWX:17}
P.~Goyal, Z.~Hu, X.~Liang, C.~Wang, and E.~Xing.
\newblock {Nonparametric Variational Auto-encoders for Hierarchical
  Representation Learning}.
\newblock {\em arXiv:1703.07027}, 2017.

\bibitem{GKATVVC:16}
I.~Gulrajani, K.~Kumar, F.~Ahmed, A.~A. Taiga, F.~Visin, D.~Vazquez, and
  A.~Courville.
\newblock {PixelVAE: A latent variable model for natural images}.
\newblock {\em arXiv:1611.05013}, 2016.

\bibitem{HJ:16}
M.~D. Hoffman and M.~J. Johnson.
\newblock {ELBO surgery: yet another way to carve up the variational evidence
  lower bound}.
\newblock {\em NIPS Workshop: Advances in Approximate Bayesian Inference},
  2016.

\bibitem{KB:14}
D.~Kingma and J.~Ba.
\newblock Adam: A method for stochastic optimization.
\newblock {\em arXiv:1412.6980}, 2014.

\bibitem{KSJCSW:16}
D.~P. Kingma, T.~Salimans, R.~Jozefowicz, X.~Chen, I.~Sutskever, and
  M.~Welling.
\newblock Improved variational inference with inverse autoregressive flow.
\newblock {\em NIPS}, pages 4743--4751, 2016.

\bibitem{KW:13}
D.~P. Kingma and M.~Welling.
\newblock Auto-encoding variational bayes.
\newblock {\em arXiv:1312.6114}, 2013.

\bibitem{LST:15}
B.~M. Lake, R.~Salakhutdinov, and J.~B. Tenenbaum.
\newblock Human-level concept learning through probabilistic program induction.
\newblock {\em Science}, 350(6266):1332--1338, 2015.

\bibitem{LM:11}
H.~Larochelle and I.~Murray.
\newblock {The Neural Autoregressive Distribution Estimator}.
\newblock {\em AISTATS}, 2011.

\bibitem{LSLW:16}
A.~B.~L. Larsen, S.~K. S{\o}nderby, H.~Larochelle, and O.~Winther.
\newblock Autoencoding beyond pixels using a learned similarity metric.
\newblock {\em ICML}, 2016.

\bibitem{LSZ:15}
Y.~Li, K.~Swersky, and R.~S. Zemel.
\newblock Generative moment matching networks.
\newblock {\em ICML}, pages 1718--1727, 2015.

\bibitem{LT:16}
Y.~Li and R.~E. Turner.
\newblock {R\'{e}nyi Divergence Variational Inference}.
\newblock {\em NIPS}, pages 1073--1081, 2016.

\bibitem{MFW:17}
L.~Maal{\o}e, M.~Fraccaro, and O.~Winther.
\newblock Semi-supervised generation with cluster-aware generative models.
\newblock {\em arXiv:1704.00637}, 2017.

\bibitem{MSJGF:15}
A.~Makhzani, J.~Shlens, N.~Jaitly, I.~Goodfellow, and B.~Frey.
\newblock Adversarial autoencoders.
\newblock {\em arXiv:1511.05644}, 2015.

\bibitem{MSCF:10}
B.~Marlin, K.~Swersky, B.~Chen, and N.~Freitas.
\newblock Inductive principles for {Restricted Boltzmann Machine} learning.
\newblock {\em AISTATS}, pages 509--516, 2010.

\bibitem{MNG:17}
L.~Mescheder, S.~Nowozin, and A.~Geiger.
\newblock {Adversarial Variational Bayes: Unifying Variational Autoencoders and
  Generative Adversarial Networks}.
\newblock In {\em ICML}, pages 2391--2400, 2017.

\bibitem{NHS:16}
E.~Nalisnick, L.~Hertel, and P.~Smyth.
\newblock {Approximate Inference for Deep Latent Gaussian Mixtures}.
\newblock {\em NIPS Workshop: Bayesian Deep Learning}, 2016.

\bibitem{NS:16}
E.~Nalisnick and P.~Smyth.
\newblock {Stick-Breaking Variational Autoencoders}.
\newblock {\em arXiv:1605.06197}, 2016.

\bibitem{RTB:16}
R.~Ranganath, D.~Tran, and D.~Blei.
\newblock Hierarchical variational models.
\newblock In {\em ICML}, pages 324--333, 2016.

\bibitem{RM:15}
D.~J. Rezende and S.~Mohamed.
\newblock Variational inference with normalizing flows.
\newblock {\em arXiv:1505.05770}, 2015.

\bibitem{RMW:14}
D.~J. Rezende, S.~Mohamed, and D.~Wierstra.
\newblock {Stochastic Backpropagation and Approximate Inference in Deep
  Generative Models}.
\newblock {\em ICML}, pages 1278--1286, 2014.

\bibitem{SM:08}
R.~Salakhutdinov and I.~Murray.
\newblock On the quantitative analysis of deep belief networks.
\newblock {\em ICML}, pages 872--879, 2008.

\bibitem{SKW:15}
T.~Salimans, D.~Kingma, and M.~Welling.
\newblock Markov chain monte carlo and variational inference: Bridging the gap.
\newblock {\em ICML}, pages 1218--1226, 2015.

\bibitem{SKTSKW:16}
C.~K. S{\o}nderby, T.~Raiko, L.~Maal{\o}e, S.~K. S{\o}nderby, and O.~Winther.
\newblock Ladder variational autoencoders.
\newblock {\em NIPS}, pages 3738--3746, 2016.

\bibitem{SC:16}
S.~Suh and S.~Choi.
\newblock {Gaussian Copula Variational Autoencoders for Mixed Data}.
\newblock {\em arXiv:1604.04960}, 2016.

\bibitem{TPB:00}
N.~Tishby, F.~C. Pereira, and W.~Bialek.
\newblock The information bottleneck method.
\newblock {\em arXiv preprint physics/0004057}, 2000.

\bibitem{TW:16}
J.~M. Tomczak and M.~Welling.
\newblock {Improving Variational Auto-Encoders using Householder Flow}.
\newblock {\em NIPS Workshop: Bayesian Deep Learning}, 2016.

\bibitem{TRB:15}
D.~Tran, R.~Ranganath, and D.~M. Blei.
\newblock {The variational Gaussian process}.
\newblock {\em arXiv:1511.06499}, 2015.

\bibitem{OKEVG:16}
A.~van~den Oord, N.~Kalchbrenner, L.~Espeholt, O.~Vinyals, A.~Graves, and
  K.~Kavukcuoglu.
\newblock {Conditional image generation with PixelCNN decoders}.
\newblock {\em NIPS}, pages 4790--4798, 2016.

\bibitem{OKK:16}
A.~van~den Oord, N.~Kalchbrenner, and K.~Kavukcuoglu.
\newblock {Pixel Recurrent Neural Networks}.
\newblock {\em ICML}, pages 1747--1756, 2016.

\bibitem{YWQSC:17}
A.~W. Yu, Q.~Lin, R.~Salakhutdinov, and J.~Carbonell.
\newblock Normalized gradient with adaptive stepsize method for deep neural
  network training.
\newblock {\em arXiv:1707.04822}, 2017.

\end{thebibliography}

\newpage
\onecolumn
\section{SUPPLEMENTARY MATERIAL}

\subsection{The gradient of Eq. \ref{eq:elbo_MC_1} and \ref{eq:elbo_MC_2} with the standard normal prior}

The gradient of $\widetilde{ \mathcal{L} }(\phi, \theta, \lambda)$ in (\ref{eq:elbo_MC_1}) and (\ref{eq:elbo_MC_2}) for the standard normal prior with respect to a single weight $\phi_{i}$ for a single data point $\mb{x}$ is the following:
\begin{align}
\frac{\partial}{\partial \phi_i} \widetilde{ \mathcal{L} }(\mb{x}; \phi, \theta, \lambda) =& \frac{1}{L} \sum_{l=1}^{L} \Big{[} \frac{1}{ p_{\theta}(\mb{x}|\mb{z}_{\phi}^{(l)})}\ \frac{\partial}{\partial \mb{z}_{\phi} } p_{\theta}(\mb{x}|\mb{z}_{\phi}^{(l)})\ \frac{\partial}{\partial \phi_i} \mb{z}_{\phi}^{(l)} -\frac{1}{q_{\phi}(\mb{z}_{\phi}^{(l)}|\mb{x})} \frac{\partial}{\partial \phi_i}q_{\phi}(\mb{z}_{\phi}^{(l)}|\mb{x}) + \label{eq:gradient_vae_1} \\
	&+ \frac{1}{p_{\lambda}(\mb{z}^{(l)}_{\phi})\ q_{\phi}(\mb{z}_{\phi}^{(l)}|\mb{x})} \Big{(} q_{\phi}(\mb{z}_{\phi}^{(l)}|\mb{x}) \frac{\partial}{\partial \mb{z}_{\phi} }p_{\lambda}(\mb{z}^{(l)}_{\phi}) - p_{\lambda}(\mb{z}^{(l)}_{\phi}) \frac{\partial}{\partial \mb{z}_{\phi} } q_{\phi}(\mb{z}_{\phi}^{(l)}|\mb{x}) \Big{)} \frac{\partial}{\partial \phi_i} \mb{z}^{(l)}_{\phi} \Big{]} .  \label{eq:gradient_vae_2}
\end{align}

\subsection{The gradient of Eq. \ref{eq:elbo_MC_1} and \ref{eq:elbo_MC_2} with the VampPrior}

The gradient of $\widetilde{ \mathcal{L} }(\phi, \theta, \lambda)$ in (\ref{eq:elbo_MC_1}) and (\ref{eq:elbo_MC_2}) for the VampPrior with respect to a single weight $\phi_{i}$ for a single data point $\mb{x}$ is the following:
\begin{align}
\frac{\partial}{\partial \phi_i} \widetilde{ \mathcal{L} }(\mb{x}; \phi, \theta, \lambda) =& \frac{1}{L} \sum_{l=1}^{L} \Big{[} \frac{1}{ p_{\theta}(\mb{x}|\mb{z}_{\phi}^{(l)})}\ \frac{\partial}{\partial \mb{z}_{\phi} } p_{\theta}(\mb{x}|\mb{z}_{\phi}^{(l)})\ \frac{\partial}{\partial \phi_i} \mb{z}_{\phi}^{(l)} + \label{eq:gradient_eep_0} \\
	&+ \frac{1}{K} \sum_{k=1}^{K} \Big{\{} \ \Big{(} \frac{q_{\phi}(\mb{z}_{\phi}^{(l)} | \mb{x})\ \frac{\partial}{\partial \phi_i} q_{\phi}(\mb{z}_{\phi}^{(l)} | \mb{u}_{k}) - q_{\phi}(\mb{z}_{\phi}^{(l)} | \mb{u}_{k})\ \frac{\partial}{\partial \phi_i} q_{\phi}(\mb{z}_{\phi}^{(l)} | \mb{x})}{ \frac{1}{K} \sum_{k=1}^{K} q_{\phi}(\mb{z}_{\phi}^{(l)} | \mb{u}_{k})\ q_{\phi}(\mb{z}_{\phi}^{(l)} | \mb{x}) } \Big{)} + \label{eq:gradient_eep_1} \\
	&+ \Big{(} \frac{ \big{(} q_{\phi}(\mb{z}_{\phi}^{(l)} | \mb{x})\ \frac{\partial}{\partial \mb{z}_{\phi} } q_{\phi}(\mb{z}_{\phi}^{(l)} | \mb{u}_{k}) - q_{\phi}(\mb{z}_{\phi}^{(l)} | \mb{u}_{k})\ \frac{\partial}{\partial \mb{z}_{\phi} } q_{\phi}(\mb{z}_{\phi}^{(l)} | \mb{x}) \big{)}\ \frac{\partial}{\partial \phi_i} \mb{z}_{\phi}^{(l)} }{ \frac{1}{K} \sum_{k=1}^{K} q_{\phi}(\mb{z}_{\phi}^{(l)} | \mb{u}_{k})\ q_{\phi}(\mb{z}_{\phi}^{(l)} | \mb{x}) } \Big{)} \Big{\}}  \Big{]} . \label{eq:gradient_eep_2} 
\end{align}

\subsection{Details on the gradient calculation in Eq. \ref{eq:gradient_eep_1} and \ref{eq:gradient_eep_2}}

Let us recall the objective function for single datapoint $\mb{x}_{*}$ using $L$ Monte Carlo sample points:
\begin{equation}\label{eq:appendix_epp_objective}
\widetilde{ \mathcal{L} }(\mb{x}_{*}; \phi, \theta, \lambda) = \frac{1}{L} \sum_{l=1}^{L} \Big{[} \ln p_{\theta}(\mb{x}_{*}|\mb{z}_{\phi}^{(l)}) \Big{]}  + \frac{1}{L} \sum_{l=1}^{L} \Big{[} \ln \frac{1}{K} \sum_{k=1}^{K} q_{\phi}(\mb{z}_{\phi}^{(l)} | \mb{u}_{k}) - \ln q_{\phi}(\mb{z}_{\phi}^{(l)}|\mb{x}_{*}) \Big{]} .
\end{equation}

We are interested in calculating gradient with respect to a single parameter $\phi_i$. We can split the gradient into two parts:
\begin{align}
\frac{\partial}{\partial \phi_i} \widetilde{ \mathcal{L} }(\mb{x}_{*}; \phi, \theta, \lambda) =& \frac{\partial}{\partial \phi_i}  \underbrace{ \frac{1}{L} \sum_{l=1}^{L} \Big{[} \ln p_{\theta}(\mb{x}_{*}|\mb{z}_{\phi}^{(l)}) \Big{]} }_{ (*) } \notag \\
	&+ \frac{\partial}{\partial \phi_i}  \underbrace{ \frac{1}{L} \sum_{l=1}^{L} \Big{[} \ln \frac{1}{K} \sum_{k=1}^{K} q_{\phi}(\mb{z}_{\phi}^{(l)} | \mb{u}_{k}) - \ln q_{\phi}(\mb{z}_{\phi}^{(l)}|\mb{x}_{*}) \Big{]} }_{ (**) } \label{eq:appendix_epp_objective_gradient}
\end{align}

Calculating the gradient separately for both $(*)$ and $(**)$ yields:
\begin{align}
\frac{\partial}{\partial \phi_i} (*) &= \frac{\partial}{\partial \phi_i} \frac{1}{L} \sum_{l=1}^{L} \Big{[} \ln p_{\theta}(\mb{x}_{*}|\mb{z}_{\phi}^{(l)}) \Big{]} \notag \\
	&= \frac{1}{L} \sum_{l=1}^{L} \frac{1}{ p_{\theta}(\mb{x}_{*}|\mb{z}_{\phi}^{(l)})}\ \frac{\partial}{\partial \mb{z}_{\phi}} p_{\theta}(\mb{x}_{*}|\mb{z}_{\phi}^{(l)})\ \frac{\partial}{\partial \phi_i} \mb{z}_{\phi}^{(l)} \label{eq:appendix_reconstruction_gradient} \\
	& \notag\\
\frac{\partial}{\partial \phi_i} (**)&= \frac{\partial}{\partial \phi_i} \frac{1}{L} \sum_{l=1}^{L} \Big{[} \ln \frac{1}{K} \sum_{k=1}^{K} q_{\phi}(\mb{z}_{\phi}^{(l)} | \mb{u}_{k}) - \ln q_{\phi}(\mb{z}_{\phi}^{(l)}|\mb{x}_{*}) \Big{]} \notag \\
	&[ \text{Short-hand notation: } q_{\phi}(\mb{z}_{\phi}^{(l)} | \mb{x}_{*}) \stackrel{\Delta}{=} q_{\phi}^{*} , \quad q_{\phi}(\mb{z}_{\phi}^{(l)} | \mb{u}_{k}) \stackrel{\Delta}{=} q_{\phi}^{k} ] \notag \\
	&= \frac{1}{L} \sum_{l=1}^{L} \Big{[} \frac{1}{ \frac{1}{K} \sum_{k=1}^{K} q_{\phi}^{k} }\ \Big{(} \frac{\partial}{\partial \phi_i} \frac{1}{K} \sum_{k=1}^{K} q_{\phi}^{k} + \frac{\partial}{\partial \mb{z}_{\phi} } \big{(} \frac{1}{K} \sum_{k=1}^{K} q_{\phi}^{k} \big{)}\ \frac{\partial}{\partial \phi_i} \mb{z}_{\phi}^{(l)}\Big{)} + \notag\\
	&\quad - \frac{1}{ q_{\phi}^{*} }\ \Big{(} \frac{\partial}{\partial \phi_i} q_{\phi}^{*} + \frac{\partial}{\partial \mb{z}_{\phi} } q_{\phi}^{*}\ \frac{\partial}{\partial \phi_i} \mb{z}_{\phi}^{(l)} \Big{)} \Big{]} \notag \\
	&= \frac{1}{L} \sum_{l=1}^{L} \Big{[} \frac{1}{ \frac{1}{K} \sum_{k=1}^{K} q_{\phi}^{k}\ q_{\phi}^{*} }\ \Big{(} \frac{1}{K} \sum_{k=1}^{K} q_{\phi}^{*}\ \frac{\partial}{\partial \phi_i} q_{\phi}^{k} + \big{(} \frac{1}{K} \sum_{k=1}^{N} q_{\phi}^{*}\ \frac{\partial}{\partial \mb{z}_{\phi} } q_{\phi}^{k} \big{)}\ \frac{\partial}{\partial \phi_i} \mb{z}_{\phi}^{(l)}\Big{)} + \notag\\
	&\quad - \frac{1}{ \frac{1}{K} \sum_{k=1}^{K} q_{\phi}^{k}\ q_{\phi}^{*} }\ \Big{(} \frac{1}{K} \sum_{k=1}^{K} q_{\phi}^{k}\ \frac{\partial}{\partial \phi_i} q_{\phi}^{*} + \frac{1}{K} \sum_{k=1}^{K} q_{\phi}^{k}\ \frac{\partial}{\partial \mb{z}_{\phi} } q_{\phi}^{*}\ \frac{\partial}{\partial \phi_i} \mb{z}_{\phi}^{(l)} \Big{)} \Big{]} \notag \\		
	&= \frac{1}{L} \sum_{l=1}^{L} \Big{[} \frac{1}{ \frac{1}{K} \sum_{k=1}^{K} q_{\phi}^{k}\ q_{\phi}^{*} }\ \frac{1}{K} \sum_{k=1}^{K} \Big{\{} \ \Big{(} q_{\phi}^{*}\ \frac{\partial}{\partial \phi_i} q_{\phi}^{k} - q_{\phi}^{k}\ \frac{\partial}{\partial \phi_i} q_{\phi}^{*} \Big{)} + \notag\\
	&\quad + \Big{(}  q_{\phi}^{*}\ \frac{\partial}{\partial \mb{z}_{\phi} } q_{\phi}^{k} - q_{\phi}^{k}\ \frac{\partial}{\partial \mb{z}_{\phi} } q_{\phi}^{*} \Big{)}\ \frac{\partial}{\partial \phi_i} \mb{z}_{\phi}^{(l)} \Big{\}}  \Big{]} \label{eq:appendix_epp_kl_gradient}
\end{align}

For comparison, the gradient of $(**)$ for a prior $p_{\lambda}(\mb{z})$ that is independent of the variational posterior is the following:
\begin{align}
& \frac{\partial}{\partial \phi_i} \Big{[} \frac{1}{L} \sum_{l=1}^{L}  \ln p_{\lambda}(\mb{z}_{\phi}^{(l)}) - \ln q_{\phi}(\mb{z}_{\phi}^{(l)}|\mb{x}_{*}) \Big{]} = \notag \\
	&[ \text{Short-hand notation: } q_{\phi}(\mb{z}_{\phi}^{(l)} | \mb{x}_{*}) \stackrel{\Delta}{=} q_{\phi}^{*} , \qquad p_{\lambda}(\mb{z}_{\phi}^{(l)}) \stackrel{\Delta}{=} p_{\lambda} ] \notag \\
	&= \frac{1}{L} \sum_{l=1}^{L} \Big{[} \frac{1}{ p_{\lambda} }\ \frac{\partial}{\partial \mb{z}_{\phi} } p_{\lambda}\ \frac{\partial}{\partial \phi_i} \mb{z}_{\phi}^{(l)} - \frac{1}{ q_{\phi}^{*} }\ \Big{(} \frac{\partial}{\partial \phi_i} q_{\phi}^{*} + \frac{\partial}{\partial \mb{z}_{\phi} } q_{\phi}^{*}\ \frac{\partial}{\partial \phi_i} \mb{z}_{\phi}^{(l)} \Big{)} \Big{]} \notag \\
	&= \frac{1}{L} \sum_{l=1}^{L} \Big{[} - \frac{1}{ q_{\phi}^{*} } \frac{\partial}{\partial \phi_i} q_{\phi}^{*} +  \frac{1}{ p_{\lambda}\ q_{\phi}^{*} }\ \Big{(} q_{\phi}^{*}\ \frac{\partial}{\partial \mb{z}_{\phi}} p_{\lambda} -  p_{\lambda}\ \frac{\partial}{\partial \mb{z}_{\phi}} q_{\phi}^{*}\Big{)}\ \frac{\partial}{\partial \phi_i}  \mb{z}_{\phi}^{(l)} \Big{]} \label{eq:appendix_standard_kl_gradient}
\end{align}

We notice that in (\ref{eq:appendix_epp_kl_gradient}) if $q_{\phi}^{*} \approx q_{\phi}^{k}$ for some $k$, then the differences $(q_{\phi}^{*}\  \frac{\partial}{\partial \phi_i} q_{\phi}^{k} - q_{\phi}^{k}\ \frac{\partial}{\partial \phi_i} q_{\phi}^{*})$ and $(q_{\phi}^{*}\ \frac{\partial}{\partial \mb{z}_{\phi} } q_{\phi}^{k} - q_{\phi}^{k}\ \frac{\partial}{\partial \mb{z}_{\phi} } q_{\phi}^{*} )$ are close to $0$. Hence, the gradient points into an average of all dissimilar pseudo-inputs contrary to the gradient of the standard normal prior in (\ref{eq:appendix_standard_kl_gradient}) that pulls always towards $\mb{0}$. As a result, the encoder is trained so that to have large variance because it is attracted by all dissimilar points and due to this fact it assigns separate regions in the latent space to each datapoint. This effect should help the decoder to decode a hidden representation to an image much easier.

\subsection{Details on experiments}

All experiments were run on NVIDIA TITAN X Pascal. The code for our models is available online at \url{https://github.com/jmtomczak/vae_vampprior}.

\subsubsection{Datasets used in the experiments}

We carried out experiments using six image datasets: static and dynamic MNIST\footnote{\url{http://yann.lecun.com/exdb/mnist/}}, OMNIGLOT\footnote{We used the pre-processed version of this dataset as in \cite{BGS:15}: \url{https://github.com/yburda/iwae/blob/master/datasets/OMNIGLOT/chardata.mat}.} \cite{LST:15}, Caltech 101 Silhouettes\footnote{We used the dataset with fixed split into training, validation and test sets: \url{https://people.cs.umass.edu/~marlin/data/caltech101_silhouettes_28_split1.mat}.} \cite{MSCF:10}, Frey Faces\footnote{\url{http://www.cs.nyu.edu/~roweis/data/frey_rawface.mat}}, and Histopathology patches \cite{TW:16}. Frey Faces contains images of size $28 \times 20$ and all other datasets contain $28 \times 28$ images. We distinguish between static MNIST with fixed binarizartion of images \cite{LM:11} and dynamic MNIST with dynamic binarization of data during training as in \cite{SM:08}.

MNIST consists of hand-written digits split into 60,000 training datapoints and 10,000 test sample points. In order to perform model selection we put aside 10,000 images from the training set. We distinguish between static MNIST with fixed binarizartion of images\footnote{\url{https://github.com/yburda/iwae/tree/master/datasets/BinaryMNIST}} \cite{LM:11} and dynamic MNIST with dynamic binarization of data during training as in \cite{SM:08}.

OMNIGLOT is a dataset containing 1,623 hand-written characters from 50 various alphabets. Each character is represented by about 20 images that makes the problem very challenging. The dataset is split into 24,345 training datapoints and 8,070 test images. We randomly pick 1,345 training examples for validation. During training we applied dynamic binarization of data similarly to dynamic MNIST.

Caltech 101 Silhouettes contains images representing silhouettes of 101 object classes. Each image is a filled, black polygon of an object on a white background. There are 4,100 training images, 2,264 validation datapoints and 2,307 test examples. The dataset is characterized by a small training sample size and many classes that makes the learning problem ambitious.

Frey Faces is a dataset of faces of a one person with different emotional expressions. The dataset consists of nearly 2,000 gray-scaled images. We randomly split them into 1,565 training images, $200$ validation images and $200$ test images. We repeated the experiment $3$ times.

Histopathology is a dataset of histopathology patches of ten different biopsies containing cancer or anemia. The dataset consists of gray-scaled images divided into 6,800 training images, 2,000 validation images and 2,000 test images.

\subsubsection{Additional results: Wall-clock times}

Using our implementation, we have calculated wall-clock times for $K=500$ (measured on MNIST) and $K=1000$ (measured on OMNIGLOT). HVAE+VampPrior was about 1.4 times slower than the standard normal prior. ConvHVAE and PixelHVAE with the VampPrior resulted in the increased training times, respectively, by a factor of $\times 1.9$/$\times 2.1$ and $\times 1.4$/$\times 1.7$ ($K=500$/$K=1000$) comparing to the standard prior. We believe that this time burden is acceptable regarding the improved generative performance resulting from the usage of the VampPrior.

\subsubsection{Additional results: Generations, reconstructions and histograms of log-likelihood}

\begin{figure}[!htbp]
\includegraphics[width=1.0\textwidth]{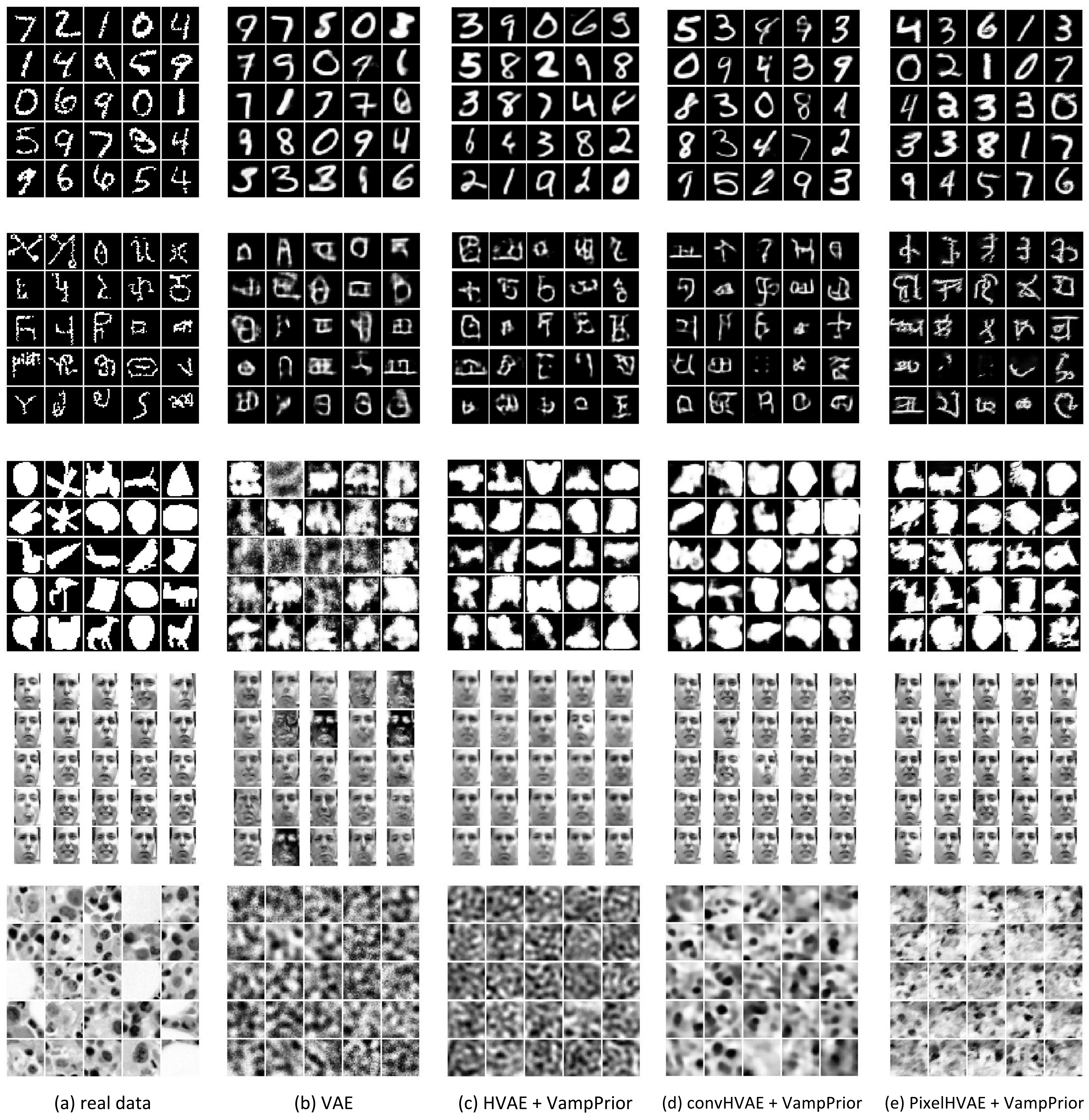}
\vskip -0.0cm
\caption{(a) Real images from test sets and images generated by (b) the vanilla VAE, (c) the HVAE ($L=2$) + VampPrior, (d) the convHVAE ($L=2$) + VampPrior and (e) the PixelHVAE ($L=2$) + VampPrior.}
\label{fig:appendix_generations}
\end{figure}

The generated images are presented in Figure \ref{fig:appendix_generations}. Images generated by HVAE ($L=2$) + VampPrior are more realistic and sharper than the ones given by the vanilla VAE. The quality of images generated by convHVAE and PixelHVAE contain much more details and better reflect variations in data.

\begin{figure}[!htbp]
\includegraphics[width=1.0\textwidth]{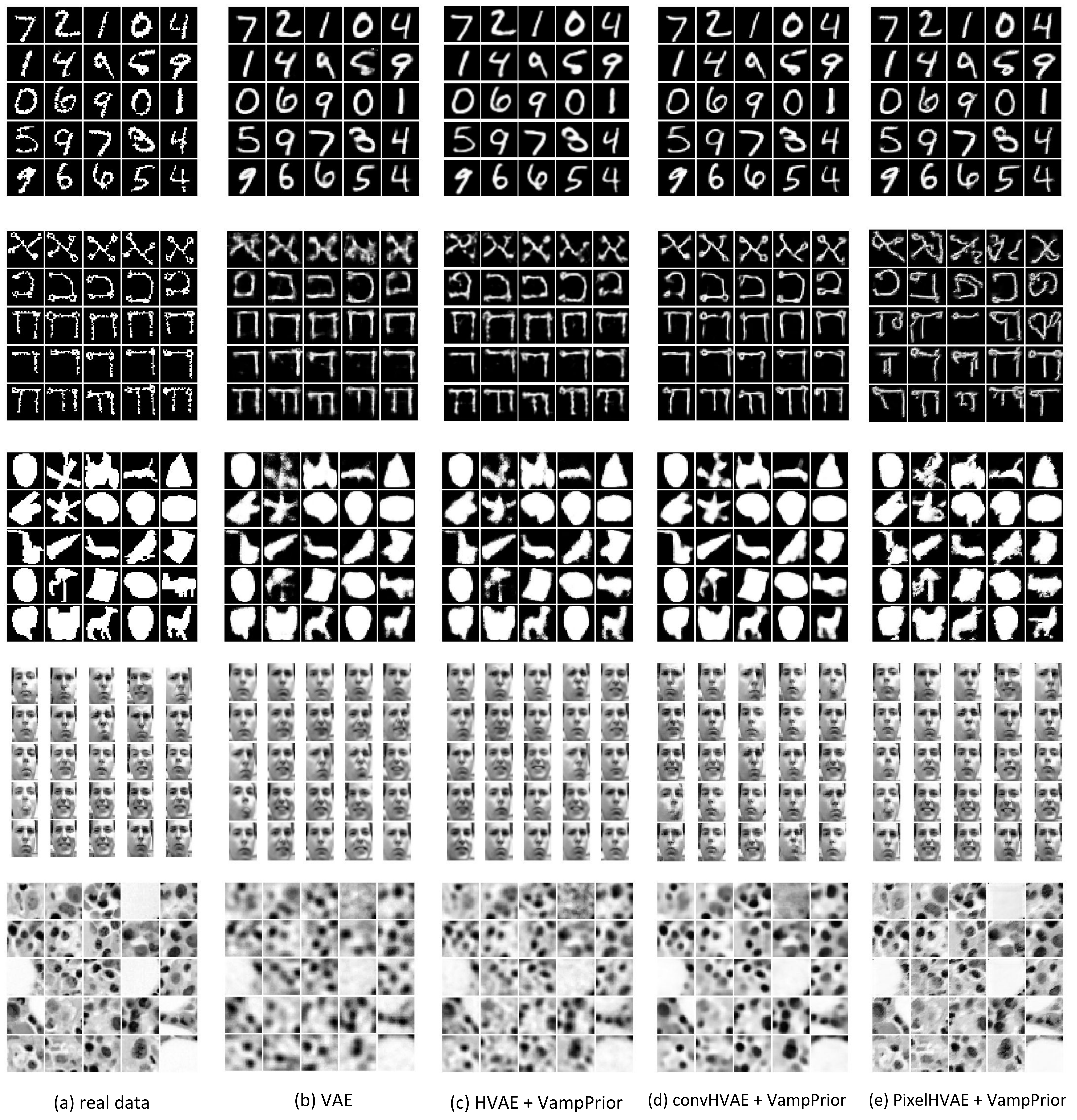}
\vskip -0.25cm
\caption{(a) Real images from test sets, (b) reconstructions given by the vanilla VAE, (c) the HVAE ($L=2$) + VampPrior, (d) the convHVAE ($L=2$) + VampPrior and (e) the PixelHVAE ($L=2$) + VampPrior.}
\label{fig:appendix_reconstructions}
\end{figure}

The reconstructions from test images are presented in Figure \ref{fig:appendix_reconstructions}. At first glance the reconstructions of VAE and HVAE ($L=2$) + VampPrior look similarly, however, our approach provides more details and the reconstructions are sharper. This is especially visible in the case of OMNIGLOT (middle row in Figure \ref{fig:appendix_reconstructions}) where VAE is incapable to reconstruct small circles while our approach does in most cases. The application of convolutional networks further improves the quality of reconstructions by providing many tiny details. Interestingly, for the PixelHVAE we can notice some "fantasizing" during reconstructing images (\textit{e.g.}, for OMNIGLOT). It means that the decoder was, to some extent, too flexible and disregarded some information included in the latent representation.

\begin{figure}[!htbp]
\includegraphics[width=0.8\textwidth]{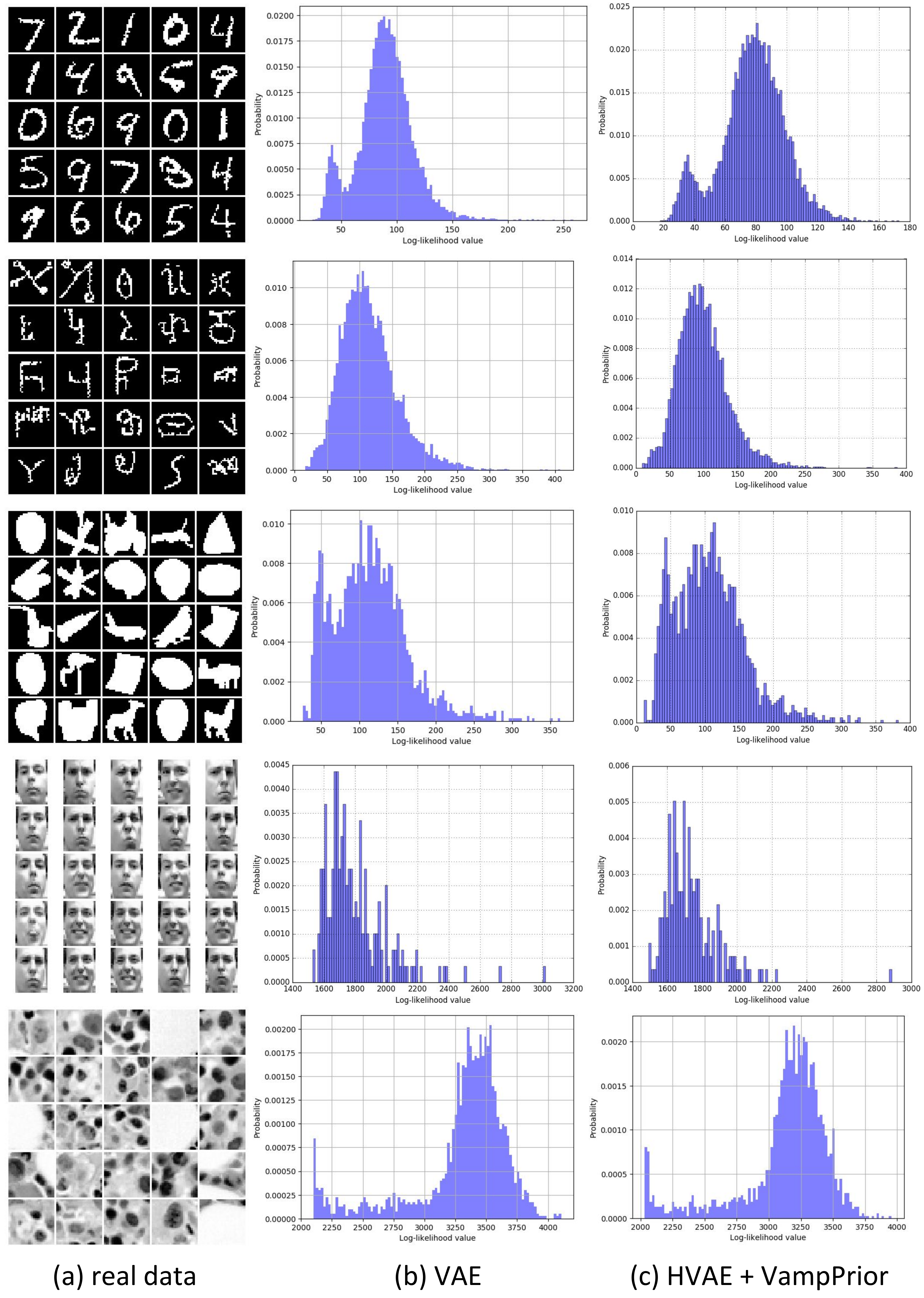}
\vskip -0.0cm
\caption{Histograms of test log-lihelihoods calculated on (from top to bottom) MNIST, OMNIGLOT, Caltech101Silhouettes, Frey Faces and Histopathology for (b) the vanilla VAE and (c) HVAE ($L=2$) + VampPrior. }
\label{fig:appendix_log_likelihoods}
\end{figure}

The histograms of the log-likelihood per test example are presented in Figure \ref{fig:appendix_log_likelihoods}. We notice that all histograms characterize a heavy-tail indicating existence of examples that are hard to represent. However, taking a closer look at the histograms for HVAE ($L=2$) + VampPrior and its convolutional-based version reveals that there are less hard examples comparing to the standard VAE. This effect is especially apparent for the convHVAE.

\end{document}